\documentclass[10pt,letterpaper]{article}

\usepackage{cogsci}
\usepackage{booktabs} 
\cogscifinalcopy 

\usepackage{multirow}
\usepackage{cmap}
\usepackage{booktabs}
\usepackage[T1]{fontenc}
\usepackage[american]{babel}
\usepackage{csquotes}
\usepackage{newtxtext,newtxmath}  
\usepackage{graphicx} 
\usepackage{subcaption}

\newcommand{\YZ}[1]{\textcolor{black}{#1}}

\usepackage[
  backend=biber,
  style=apa,
  natbib=true,
  eprint=true,
  annotation=false,
]{biblatex}
\usepackage{float} 

\usepackage[hidelinks]{hyperref}
\usepackage{xcolor}

\definecolor{OliveGreen}{rgb}{0,0.6,0}

\title{Modeling Human-Like Color Naming Behavior in Context} 

\author[1]{\mbox{Yuqing Zhang}}
\author[2]{\mbox{Ecesu Ürker}}
\author[3]{\mbox{Tessa Verhoef}}
\author[4]{\mbox{Gemma Boleda}}
\author[1]{\mbox{Arianna Bisazza}}
\affil[1]{Center for Language and Cognition, University of Groningen}
\affil[2]{Department of Translation and Language Sciences, Universitat Pompeu Fabra}
\affil[3]{Leiden Institute of Advanced Computer Science,  Leiden University}
\affil[4]{Department of Translation and Language Sciences, Universitat Pompeu Fabra / ICREA}

\begin{document}

\maketitle

\begin{abstract} 
Modeling the emergence of human-like lexicons in computational systems has advanced through the use of interacting neural agents, which simulate both learning and communicative pressures. The NeLLCom-Lex framework \citep{zhang-etal-2025-nellcom} allows neural agents to develop pragmatic color naming behavior
and human-like lexicons through supervised learning (SL) from human data and reinforcement learning (RL) in referential games. Despite these successes, the lexicons that emerge diverge systematically from human color categories, producing highly non-convex regions in color space, which contrast with the convexity typical of human categories. To address this, we introduce two factors, upsampling rare color terms during SL and multi-listener RL interactions, and adopt a convexity measure to quantify geometric coherence. We find that upsampling improves lexical diversity and system-level informativeness of the color lexicon, while many-listener setups promote more convex color categories. The combination of moderate upsampling and multiple listeners produces lexicons most similar to human systems.

\end{abstract}

\section{Introduction}

The relationship between the use of a language and its properties as a linguistic system is highly dynamic, such that language use shapes language systems and vice versa~\citep{clark1996using,hawkins2004efficiency,beckner2009language, brochhagen2022languages,regier2015word}.
Studying this interaction is challenging, 
as it emerges from system-level dynamics among large populations 
and unfolds over extended historical periods~\citep{hopper2003grammaticalization, campbell2013historical}. 
Computational modeling offers a powerful approach, enabling simulation of system-level pressures in a controlled way while varying parameters inaccessible in experiments with humans or field work \citep{steels1997synthetic, steels1998spatially, cangelosi2002computer, de2006computer,brochhagen2018signaling}. Recent advances in deep learning have substantially expanded what is possible in this domain through powerful and flexible neural-agent models \citep{kharitonov-etal-2019-egg, lazaridou2020emergent, chaabouni2021communicating, lian2023communication}.

\begin{figure}[t]
    \centering
    \includegraphics[width=\linewidth]{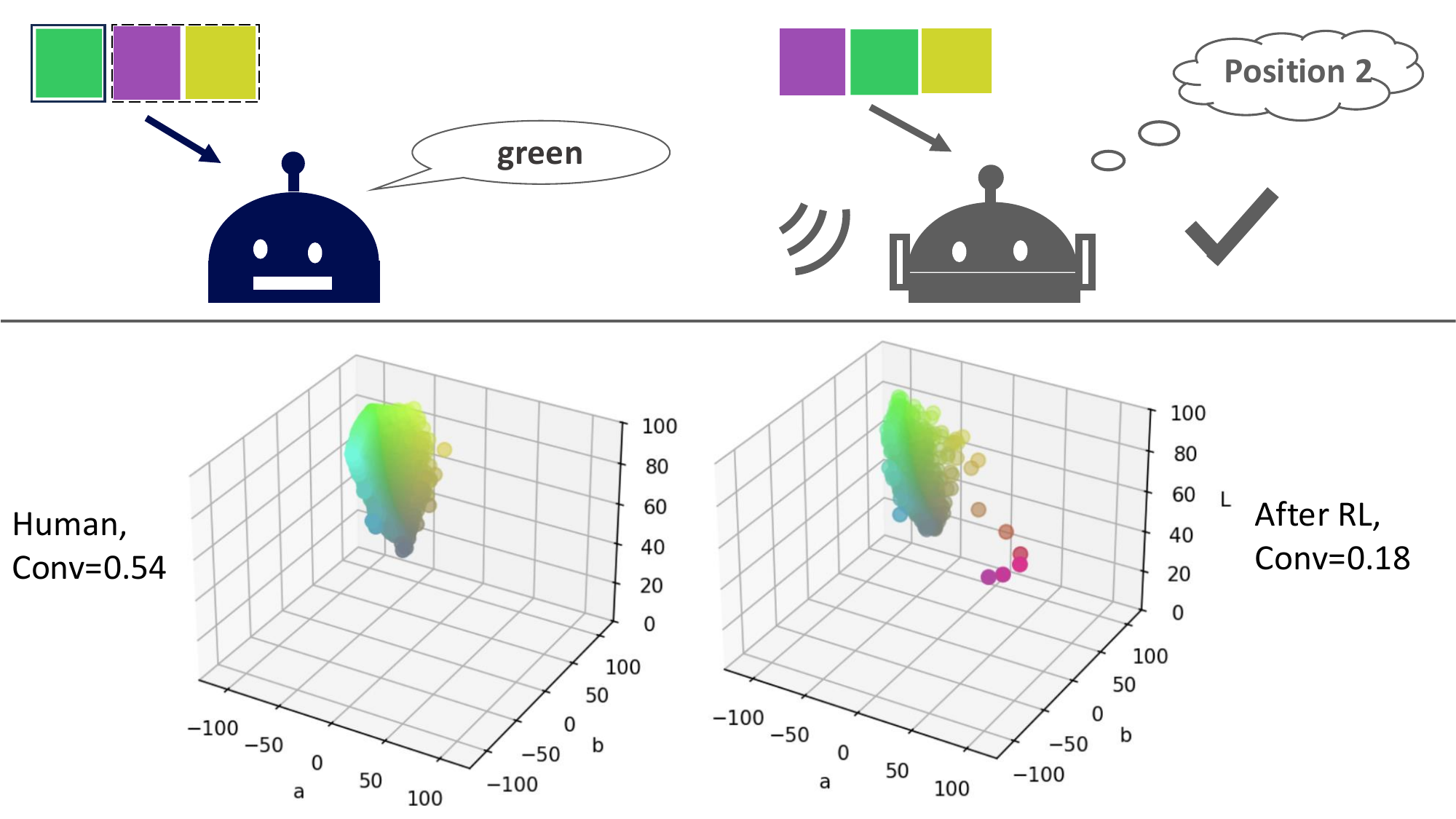} 
    \caption{Agents develop non-convex color representations as a result of playing referential games. The top row illustrates the referential game setup, in which a target color chip (outlined with a solid line) is presented among distractors (dashed line). The bottom row compares human and agent color representations for the color term \textit{green}. 
    }
    \label{fig:color_context}
\end{figure}

In this paper, we use this method to investigate the relationship between language use and language system in the domain of the lexicon. 
We adopt the recently introduced NeLLCom-Lex framework \citep{zhang-etal-2025-nellcom}, aimed at analyzing how communicative factors shape lexicons by isolating interactional pressures from population-level transmission dynamics. After first training agents on human data from a color discrimination task, we study how communicative use reshapes an already grounded lexicon. 
In this framework, agents are first pretrained via SL and subsequently interact through RL in referential games, as illustrated in Figure~\ref{fig:color_context} (top). During these interactions, the speaker and the listener agents are presented with a target and two distractor color chips. The speaker generates a color term to convey the target to the listener (e.g. \textit{green} in this example) and the listener agent selects a color chip based on the speaker's message.
In \citet{zhang-etal-2025-nellcom}, agents trained in this framework were shown to develop context-sensitive pragmatic behavior. 

Despite these successes, 
the lexicons that emerge from NeLLCom-Lex agents show systematic divergences from human lexicons; in particular, they establish color categories that are often highly non-convex in color space, and agents seem to come up with partner-specific "conceptual pacts" \citep{brennan1996conceptual} (see Figure~\ref{fig:color_context}, bottom, for an illustration). 
Although such systems can support effective communication among agents, these properties contrast with well-established characteristics of human color categories, which typically correspond to convex regions in perceptual space \citep{gardenfors2014geometry, jager2007evolution, jager2010natural}. This constraint, known as the \textit{convexity universal}, states that the denotation of every color term is a convex region in the CIELAB color space that approximates human color vision \citep{brainard2003color}. \citet{gardenfors2000conceptual, gardenfors2014geometry} has argued more generally that convexity is a hallmark of human categories, and hence of word meaning.

If we are to use neural agents to model the dynamics between word use and lexical systems, we need to ensure that their lexicons exhibit key properties of human lexicons. Since convexity is one of the necessary characteristics of human lexical categories \citep{carlsson2024cultural}, 
this motivates a closer examination of the conditions that give rise to non-convex denotations in the first place.
In this paper, we investigate factors that are crucial for convexity to arise, and focus on two factors that shape the geometry of emergent lexicons but were not central in prior analyses: sparsity of exposure to some color terms, and listener diversity. 
The listener-diversity manipulation is motivated by the hypothesis that greater interlocutor variety makes conceptual pacts more difficult to establish, thereby promoting more convex semantic categories. An alternative explanation, however, is that non-convexity arises from limited exposure to certain meanings, resulting in poorly learned denotations. Although this account is unlikely given that even high-frequency terms exhibit non-convexity. We therefore introduce upsampling to control for data sparsity, allowing us to disentangle the effects of listener diversity from those of general learning difficulty.
Crucially, we also track other characteristics of human lexicons that have been described in the literature, including vocabulary size and system informativeness \citep{regier2015word,gualdoni-boleda-2024-objects}. This allows us to examine how convexity relates to these properties: for example, a one-word lexical system would be fully convex. 
We find that upsampling rare color terms during SL increases lexical diversity and informativeness while reducing semantic drift, whereas interacting with many listeners constrains vocabulary size and informativeness but promotes a more convex color space. The combination of moderate upsampling and a many-listener setup produces lexicons that are most human-like across multiple metrics, highlighting the importance of data exposure and population structure in shaping emergent lexical systems\footnote{Code and materials are available at \url{https://github.com/yuqing0304/NeLLCom_Lex_CogSci}.}.

\section{Related Work}

\textbf{Emergent Communication in the Color Domain}


Drawing on the World Color Survey (WCS) 
\citep{berlin1969basic,cook2005world}, the color domain has become a widely used test case for agent-based models of language emergence. 
Early studies demonstrated that artificial agents \citep{steels2005coordinating} and even embodied robots \citep{bleys2009grounded} could develop color categories from scratch through repeated language games.
Subsequent work using similar language game paradigms produced increasingly human-like color naming systems by incorporating simple perceptual constraints \citep{Baronchelli2010, loreto2012origin}. Comparable results have been achieved with more recent neural deep-learning agents that learn from interactions through RL \citep{carlsson2024cultural,Carlsson2020, chaabouni2021communicating,tucker2025towards}, building on the information-theoretic model of \citet{zaslavsky2018efficient}. 
In these set-ups, agents develop their own color naming systems from scratch. In contrast, NeLLCom-Lex \citep{zhang-etal-2025-nellcom} 
initializes agents with a human color naming system to enable more interpretable analyses, and allow for more focused experiments on realistic scenarios of semantic change. 

\paragraph{Color Naming in Context}
Prior approaches mainly model color lexicons as a partition of the perceptual color space, with a unique label for each color region, abstracting away from how speakers select among alternative terms in specific communicative situations. Natural languages typically provide multiple alternative labels for the same referent and a single color, for example, can be described as either purple or magenta \citep{gualdoni-boleda-2024-objects}. Recently, pragmatic modeling work employed color reference games to investigate agent color naming behavior in context. For instance, \citet{monroe-etal-2017-colors} model speakers and listeners using the Rational Speech Acts framework and show that speakers produce longer, more descriptive utterances in more difficult contexts. Similarly, \citet{white2020learning} demonstrate that agents trained to approximate pragmatic reasoning objectives adjust utterance length in response to communicative context. 
 In contrast to these studies, NeLLCom-Lex does not rely on multi-word utterances or utterance length as a measure of effort, but rather accounts for variation in the informativeness of individual word choices, by adopting a measure developed by \citet{gualdoni-boleda-2024-objects} that takes into account both context adaptation in language use and the structure of lexical systems. NeLLCom-Lex demonstrated that access to contextual information during RL is important for developing human-like pragmatic behavior. 
Our work builds directly on this framework, focusing on the geometric and semantic properties of the emergent lexicons.

\paragraph{Geometry of Meaning Spaces}
The geometric organization of lexical categories has long been a topic of interest in cognitive semantics. Conceptual space theory argues that natural language categories tend to form approximately convex regions in perceptual and conceptual spaces \citep{gardenfors2014geometry}. Empirical studies of color naming across languages support this view, showing that color categories are typically compact and contiguous in perceptual color space.
Recent computational work has formalized convexity as a measurable property of semantic systems. \citet{steinert2020ease} show that color systems with more convex categories are easier for neural networks to learn, providing a learning-based explanation for semantic universals. Their convexity metric, based on the ratio between category size and its convex hull, offers a principled way to quantify the geometric coherence of lexical categories and has been used by \citet{carlsson2024cultural} to study the pressures underlying the emergence of efficient, human-like color naming systems. Building on this literature, we adopt this metric to analyze the convexity of the color categories acquired and developed by our agents.

\paragraph{Frequency Effects in Learning}
Frequency plays a central role in the learning process across both humans and computational models. In human language acquisition, high-frequency words are typically learned earlier, show greater semantic stability, and are more resistant to attrition \citep{ellis2002frequency}.
Similar biases arise in computational models trained on imbalanced data. Neural networks are known to overrepresent frequent categories while under-differentiating rare ones, a problem widely studied under class imbalance \citep{henning2023survey}. Data-level interventions such as upsampling minority classes have been shown to partially mitigate these biases by increasing effective exposure during training \citep{chawla2002smote}.

\paragraph{Listener Diversity}
Most emergent communication models focus on a single speaker-listener pair, which is also the case in the original NeLLCom-Lex. 
This setup allows lexical conventions to overfit to a particular partner, potentially encouraging idiosyncratic and non-convex category boundaries. While there is robust evidence for partner-specific accommodation in human communication \citep{brennan1996conceptual, pickering2004toward}, research also suggests that when partner-specificity offers no communicative advantage, speakers converge on partner-independent, aggregate statistics across their interlocutors \citep{ostrand2019repeat}. Thus, communicating with diverse listeners may favor more systematic and listener-agnostic linguistic systems, as also demonstrated in laboratory experiments where larger groups of participants developed more systematic and compositional lexicons than smaller groups \citep{raviv2019larger}. 
Agent-based simulations similarly show that when a speaker interacts with multiple listeners sequentially, languages that are more compositional and easier to teach to newcomers emerge \citep{li2019ease}. In addition, representations learned in populations are less idiosyncratic and more shareable than those learned in fixed pairs \citep{tieleman2019shaping}, and promote compositionality \citep{michelrevisiting}. Here, we therefore introduce a \textit{many-listeners setup} in NeLLCom-Lex, in which one speaker communicates with multiple listeners to test how it affects convexity.


\section{Experimental Setup}

\YZ{In the NeLLCom-Lex framework \citep{zhang-etal-2025-nellcom}, both speaker and listener are implemented as feed-forward neural networks operating on CIELAB color representations. The speaker encodes the target and its context (when context-aware) and produces a discrete message via a classifier, while the listener selects a referent by comparing the message embedding with candidate color embeddings. During SL, agents are trained on human-produced labels using cross-entropy loss, whereas during RL they are jointly optimized based on communicative success in referential games.
Building on this framework, we introduce two main changes to test their effects on the properties of the resulting lexicon.}

In our new \textit{many-listeners setup} one speaker communicates with 5 or 30 listeners\footnote{In NeLLCom-Lex terms, a speaker (or listener) refers to a randomly initialized speaker (listener) network.
Multiple speakers (listeners) then refer to multiple networks initialized with different random seeds prior to training.} , 
each of which is an independently initialized listener network trained for 30 SL epochs on the human data.
During the subsequent 30-epoch RL phase, referential game training is distributed across listeners for controlled comparison. \YZ{Specifically, listeners are randomly ordered without replacement, and the speaker interacts with each listener for a fixed number of consecutive RL epochs before moving to the next (e.g., with 5 listeners, each listener is trained for 6 consecutive RL epochs).}

Second, severe frequency imbalance in the SL data, \YZ{with a mean frequency of 328 and a maximum frequency of 2585,} biases the initial lexical representations learned by the speaker. High-frequency color terms receive strong grounding, whereas rare colors are weakly represented. In addition, RL tends to amplify initially successful mappings, causing rare terms to be forgotten or absorbed into broader categories. To address this issue, we introduce an \textit{upsampling setup} where rare color samples are duplicated during SL until each reaches a target count $N$,
while leaving the RL phase unchanged\footnote{\YZ{More complex augmentation strategies, such as slightly corrupting or modifying existing rare color samples to generate color variants, are an interesting direction, which we leave for future work.}}. 
\YZ{We set the target upsampling levels $N$ to 100 and 200 to represent moderate and strong rebalancing of the highly skewed frequency distribution. These values increase exposure to rare terms while still preserving relative frequency differences.}

Training follows \citet{zhang-etal-2025-nellcom}'s SL$+$RL$+$ pipeline (SL with context followed by RL with context), which yielded the lexicon with the most human-like properties. All architectural parameters and hyperparameters are kept identical.\footnote{We increase the granularity of the color space by representing CIELAB values with one decimal place instead of integers. We also note that \citet{zhang-etal-2025-nellcom}'s color sampling excludes greyish colors, which may limit generalization to pale shades; accordingly, we adjust the sampling criteria to include all saturation levels.}

\subsection{Datasets}
\label{sec:dataset}
For SL, we use the English \textbf{Colors} dataset processed by \citet{gualdoni-boleda-2024-objects}, originally collected by \citet{monroe-etal-2017-colors}. 
The data come from a dyadic reference game in which a speaker describes a target color chip among two distractors, and the listener needs to select the correct target.
Context difficulty depends on visual similarity: \textsc{\textbf{far}} (both distractors distinct), \textsc{\textbf{split}} (one distinct, one similar), and \textsc{\textbf{close}} (both similar). We use all successful single-word trials, containing 15,434 instances (9,309 far, 3,886 split, 2,239 close). For SL, 3K instances are held out for testing (\textsc{test}$_{hum}$), and the remaining 12.4K are used for training (\textsc{train}$_{hum}$), similar to \citet{zhang-etal-2025-nellcom}'s original setup.
For RL, where human labels are not required, we generate synthetic far/close color triplets following the sampling procedure of \citet{monroe-etal-2017-colors}. Both RL training and evaluation are conducted on generated data whose context distribution matches that of the human \textbf{Colors} dataset used for SL. In total, we use 12.4K generated triplets for training (\textsc{train}$_{gen,distH}$) and 15.4K for evaluation (\textsc{test}$_{gen,distH}$).

\subsection{Evaluation}


Following \citet{zhang-etal-2025-nellcom}, agent performance is evaluated using communication accuracy and pragmatic adaptation, while properties of the emergent lexicon are characterized using several system-level measures. In addition, we introduce convexity as a key metric to more closely examine the geometric structure of lexical categories.

\vspace{.1cm}\noindent\textbf{Communication accuracy} ($Acc_{comm}$) is defined as the proportion of trials in which the listener correctly identifies the target color given the input from the speaker.

\vspace{.1cm}\noindent\textbf{Pragmatic adaptation} is assessed by testing whether agents adapt their naming choices to task difficulty, quantified by the link between word informativeness and context ease. 
Word informativeness ($I_w$) is computed from the spread of a word’s denotation in CIELAB space \citep{gualdoni-boleda-2024-objects}. For a color term $w$, spread ($S_w$) is defined as the average pairwise Euclidean distance between all color chips denoted by $w$, that is
$S_w = (\sum_i \sum_{j \neq i} d(o_i, o_j))/P$, where $P$ is the number of unique object pairs and $d(o_i, o_j)$ is the Euclidean distance between objects $o_i$ and $o_j$. Informativeness is defined as $I_w = 1 / S_w$, such that more compact denotations correspond to higher $I_w$.
%
Context ease ($E_{ctx}$) refers to the distance between the target color chip and the hardest distractor in the context.
A linear mixed-effects model is fitted to predict the informativeness of the produced word ($I_w$) from $E_{ctx}$, with agent seeds and target chips as random effects.
For each setting, we report $\beta(E_{ctx})$, the estimated effect of context
ease on word informativeness.

Our system-level metrics are also taken from \citet{zhang-etal-2025-nellcom}:
\textbf{System-level informativeness} \citep{gualdoni-boleda-2024-objects} is defined as the average informativeness $I_w$ of the words used to solve \emph{N} interactions,
$I_L= (\sum_{i=1}^NI_w^i)/N$.
Human-like lexicons exhibit intermediate $I_L$ values, supporting meaningful distinctions without being overly specific \citep{gualdoni-boleda-2024-objects}.
\textbf{Lexical diversity} is defined as the number of different color words used by the speaker. 
\textbf{Semantic drift} quantifies how agents' lexicons diverge from the human lexicon by computing the Euclidean distance between their respective prototypes for each word \citep{gualdoni_whats_2023}. 

In addition to these measures, we introduce a \textbf{convexity} measure to examine the geometric structure of the agents’ conceptual space. Specifically, we adopt the procedure proposed by \citet{steinert2020ease}, whereby the degree of convexity of a color system is defined as its proximity to the closest convex color system. Following this approach, for each color term $c$ in a lexicon $L$, we compute the convex hull of $c$, defined as the smallest convex region that contains all CIELAB points associated with $c$. We then calculate the proportion of this convex hull that is already covered by $c$, which constitutes the convexity degree of $c$. To obtain the overall degree of convexity of $L$, we deviate from the original measure in \citet{steinert2020ease} by averaging the convexity scores of individual color terms without applying any weights\footnote{We use an unweighted average because our production data contain many high-frequency general terms, which would otherwise dominate the score and obscure rarer specific terms.}:
\begin{equation} 
\text{Convexity(\textit{L})} = \frac{1}{T}\sum_{i=1}^T\frac{\lvert\textit{c}^i\lvert}{\lvert\text{ConvexHull}(c^i)\lvert} 
\end{equation}
where $T$ denotes the number of color terms in the lexicon.

All experiments are repeated with 10 random seeds, and agent production is evaluated after 30 epochs of RL training.

\begin{table*}[h]
\centering
\small
\setlength{\tabcolsep}{6pt}
\caption{Agents' production properties 
under different listener setups and upsampling strategies. Numbers show the mean across 10 seeds.
$\beta(E_{ctx})$: effect of context ease on word informativeness.
$|W|$: lexical diversity; $I_L$: system-level informativeness; $D_L$: semantic drift.
Human values are computed using the full 15.4K \textbf{Colors} dataset \citep{gualdoni-boleda-2024-objects}.}
\label{tab:main}
\begin{tabular}{lcc ccc ccc c}
\toprule
\multirow{2}{*}{Conditions} 
& \multicolumn{2}{c}{Setup} 
& \multicolumn{2}{c}{Communication} 
& \multicolumn{3}{c}{Lexicon Properties} 
& \multirow{2}{*}{$D_L$} \\
\cmidrule(lr){2-3}
\cmidrule(lr){4-5}
\cmidrule(lr){6-8}
& Listeners 
& Upsampling 
& $Acc_{comm}$
& $\beta(E_{ctx})$  
& $|W|$ 
& $I_L$ 
& Convexity 
& \\
\midrule
\multirow{3}{*}{\textbf{SL}} 
& N/A & 0   & 0.87 & -0.008 & 13.6 & 3.18 & 0.60 & 10.31 \\
& N/A & 100 & 0.86 & -0.010 & 28.6 & 3.21 & 0.60 & 12.24 \\
& N/A & 200 & 0.87 & -0.013 & 35.9 & 3.36 & 0.42 & 12.15 \\
\midrule
\multirow{9}{*}{\textbf{SL$+$RL}} 
& 1  & 0   & 0.94 & -0.004 & 31.8 & 2.62 & 0.24 & 39.69 \\
& 1  & 100 & 0.94 & -0.004 & 39.0 & 2.68 & 0.22 & 35.11 \\
& 1  & 200 & 0.94 & -0.004 & 42.6 & 2.74 & 0.20 & 27.11 \\
& 5  & 0   & 0.93 & -0.005 & 23.6 & 2.48 & 0.28 & 41.98 \\
& 5  & 100 & 0.93 & -0.005 & 31.4 & 2.49 & 0.30 & 37.42 \\
& 5  & 200 & 0.93 & -0.005 & 37.6 & 2.54 & 0.26 & 28.51 \\
& 30 & 0   & 0.92 & -0.005 & 21.7 & 2.38 & 0.31 & 49.19 \\
& 30 & 100 & 0.92 & -0.005 & 29.7 & 2.39 & 0.37 & 37.72 \\
& 30 & 200 & 0.93 & -0.006 & 37.2 & 2.48 & 0.26 & 27.74 \\
\midrule
\textbf{Human} 
& -- & -- & 1.00 & -0.008 & 49.0 & 2.78 & 0.32 & -- \\
\bottomrule
\end{tabular}
\end{table*}

\begin{figure*}[h]
\centering
\includegraphics[width=\textwidth]{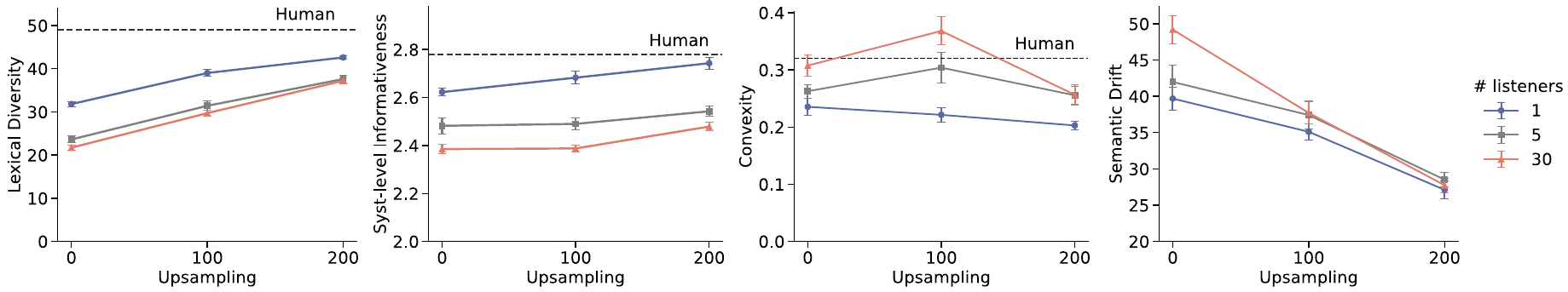}
\caption{Lexicon properties across listener setups and upsampling levels. Each marker represents the mean across 10 seeds, with bars showing standard error. 
The human reference values are indicated by dashed horizontal lines.}
\label{fig:lexicon_properties}
\end{figure*}

\section{Results and Discussion}

Results are reported in Table~\ref{tab:main}. 
Overall, agents achieve high communication accuracy across conditions, \YZ{comparable to the accuracy levels reported by \citet{monroe-etal-2017-colors} (far: 97\%, split: 90\%, close: 83\%)}. 
However, accuracies below 90\% before RL indicate that SL alone does not fully support robust communication on the test set.
Incorporating RL further improves accuracy, suggesting that communication-based training facilitates more effective language use, and confirming the findings of \citet{zhang-etal-2025-nellcom}. 
With respect to the newly added factors, we observe that accuracy is slightly lower when speakers interact with a larger number of listeners, but remains stable under increased upsampling for both SL and RL. 

As for pragmatic naming behavior, neural agents exhibit context-sensitive pragmatic adaptation under all conditions (all $\beta(E_{ctx})$ values are significant, with $p$$<$0.001, and standard errors for all estimates below 0.001). In addition, pragmatic adaptation is already strong after SL and slightly exceeds that observed after RL, consistently with \citet{zhang-etal-2025-nellcom}. 

Having assessed that agents broadly succeed at context-sensitive language use, we turn to the properties of the learned lexicons, shown in Table~\ref{tab:main} and illustrated in Figure~\ref{fig:lexicon_properties}, focusing on lexical diversity, informativeness, and convexity.

\vspace{.1cm}\noindent\textbf{Lexical Diversity. }
Comparing SL and SL$+$RL, we observe that RL training overall increases lexical diversity, replicating the patterns in \citet{zhang-etal-2025-nellcom}.
The number of listeners affects lexical diversity. With no upsampling, $|W|$ decreases with fewer listeners. Even with higher upsampling, many-listener setups still limit vocabulary usage compared to few-listener setups. This suggests that interacting with multiple listeners constrains the agents to use a more consistent and focused set of words.
%
For a fixed listener setup, upsampling consistently increases lexical diversity. For instance, in the SL-only condition, $|W|$ increases with higher levels of upsampling. Similarly, in the SL$+$RL condition with one listener, $|W|$ increases across the same upsampling levels. 
This indicates that upsampling allows the agents to expand their vocabulary and use a wider range of words.
Overall, the maximum lexical diversity is achieved with a single listener and the highest level of upsampling, yielding the richest vocabulary.

\vspace{.1cm}\noindent\textbf{System-level Informativeness.} Comparing SL and SL$+$RL, RL training is associated with lower $I_L$, consistent with the patterns reported by \citet{zhang-etal-2025-nellcom}, suggesting that SL-trained agents may develop lexicons that encode redundant distinctions, and the emergent lexicon gets optimized through interactive communication.
%
The number of listeners affects $I_L$. Similar to lexical diversity, $I_L$ decreases as the number of listeners increases, regardless of the upsampling level. This likely reflects the fact that, with a fixed number of interactions, increasing the number of listeners makes it more difficult for each agent to learn a well-formed lexicon. As a result, communicating with multiple listeners favors more general and conservative lexical choices, leading to lower diversity and $I_L$.
%
For a fixed listener setup, upsampling leads to a moderate but consistent increase in informativeness.
In the SL-only condition, $I_L$ rises as upsampling increases.
A similar trend is observed in the SL$+$RL condition with one listener, where $I_L$ increases for the same upsampling settings.
This suggests that exposure to more balanced training data enables agents to construct lexicons that convey meanings more informatively.
Overall, system-level informativeness is most human-like in the SL$+$RL setting with a single listener and the highest upsampling level. These results suggest that optimizing input for training 
can shift agents away from overly fine-grained lexical distinctions toward more efficient lexicons, bringing the agents’ informativeness closer to the human level. 

\vspace{.1cm}\noindent\textbf{Convexity.} SL-trained agents exhibit very high convexity across conditions, suggesting that they acquire structured semantic representations, likely due to low lexical diversity. In contrast, RL training reduces convexity to levels closer to or slightly below those observed in humans.
%
Interacting with more listeners consistently increases convexity.
Under SL$+$RL with no upsampling, convexity rises from 0.24 (1 listener) to 0.28 (5 listeners) and 0.31 (30 listeners), a trend that persists under higher upsampling levels. Thus, interacting with multiple listeners may encourage agents to adopt more compact and consistent color representations. 
%
The effect of upsampling on convexity is non-monotonic. Since upsampling tends to increase lexical diversity, excessive upsampling makes convexity harder to maintain, given that lexicons learned after SL alone are already highly convex.
Consistent with this intuition, upsampling-200 setting leads to a decline in convexity across conditions. For instance, in SL, convexity drops from 0.60 to 0.42 as upsampling increases from 0 to 200, alongside a vocabulary increase from $|W|$=13.6 to 35.9. A similar downward trend is observed in SL$+$RL.
However, intriguingly, moderate upsampling improves convexity under many-listener setups. 
For example, in the SL$+$RL condition with five listeners, convexity increases from 0.28 to 0.30 at an upsampling level of 100, but decreases to 0.26 when upsampling is further increased to 200.
This pattern, also visible in the general trend in Figure~\ref{fig:lexicon_properties}, suggests a potential Goldilocks principle\footnote{Establishing a strict Goldilocks optimum would require a more systematic exploration of intermediate upsampling values, which we leave for future work.}, named after the fairy tale Goldilocks and the Three Bears, in which the optimal choice is "just right" rather than too extreme \citep{brochhagen2022languages}. In our case, moderate upsampling yields a medium-sized effective lexicon (around 30 words), and multiple interlocutors ensure convexity, together helping agents learn coherent partitions of the color space. In contrast, excessive upsampling yields a larger-sized lexicon (around 37 words) that likely introduces finer-grained distinctions that may fragment semantic regions, making it more difficult to maintain high convexity. 
Notably, SL$+$RL settings with 30 listeners and moderate upsampling level (i.e., 100) achieve \YZ{the highest convexity across conditions and, though not closest to the human value, provide the best overall balance across all metrics,} suggesting that interacting with many listeners combined with moderate data augmentation promotes more human-like semantic organization. Such an effect is consistent with prior work showing that increasing the number of interacting agents reduces idiosyncratic co-adaptation \citep{graesser2019emergent} and aligns with findings that greater input variability in groups can drive languages toward simpler and more systematic systems in human experiments  \citep{raviv2019larger}.

\begin{figure}[t]
\centering
\begin{subfigure}[b]{0.32\linewidth}
  \centering
  \includegraphics[width=\linewidth]{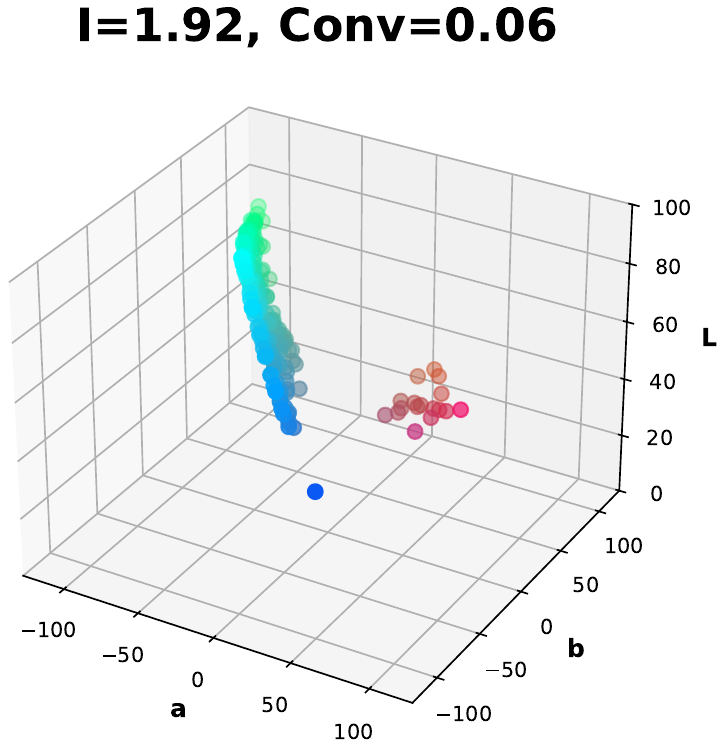}
\end{subfigure}
\begin{subfigure}[b]{0.32\linewidth}
  \centering
  \includegraphics[width=\linewidth]{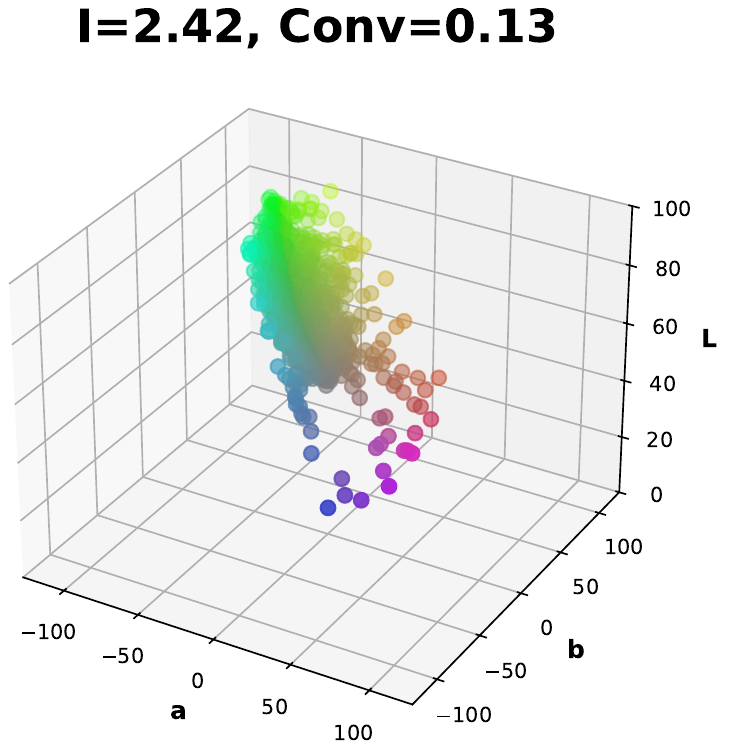}
\end{subfigure}
\begin{subfigure}[b]{0.32\linewidth}
  \centering
  \includegraphics[width=\linewidth]{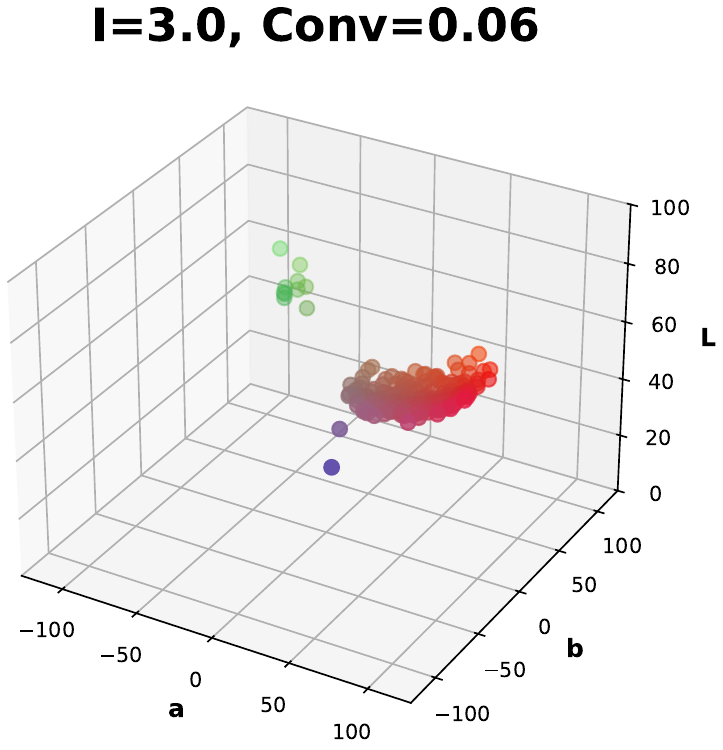}
\end{subfigure}
\vspace{0.5em}
\begin{subfigure}[b]{0.32\linewidth}
  \centering
  \includegraphics[width=\linewidth]{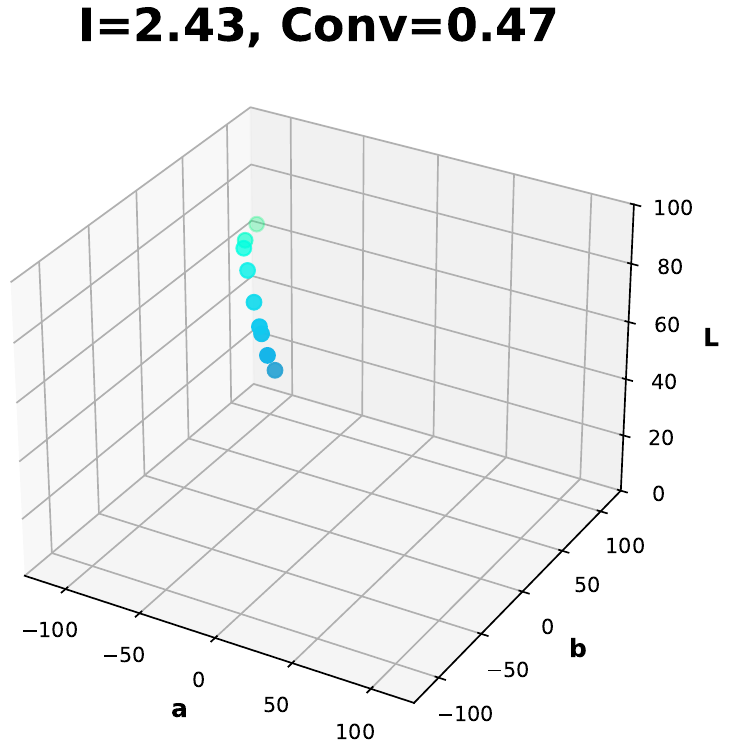}
  \caption{\textit{aqua}}
\end{subfigure}
\begin{subfigure}[b]{0.32\linewidth}
  \centering
  \includegraphics[width=\linewidth]{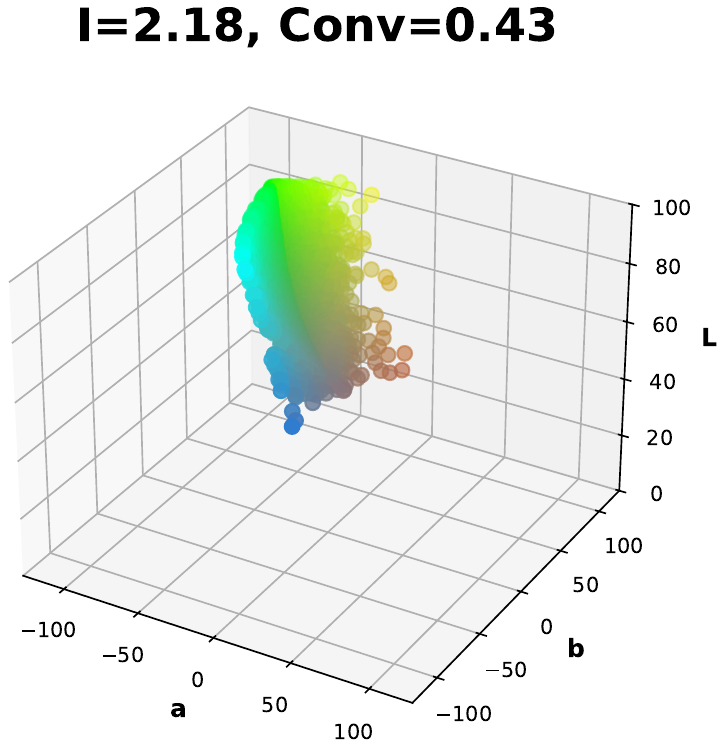}
  \caption{\textit{green}}
\end{subfigure}
\begin{subfigure}[b]{0.32\linewidth}
  \centering
  \includegraphics[width=\linewidth]{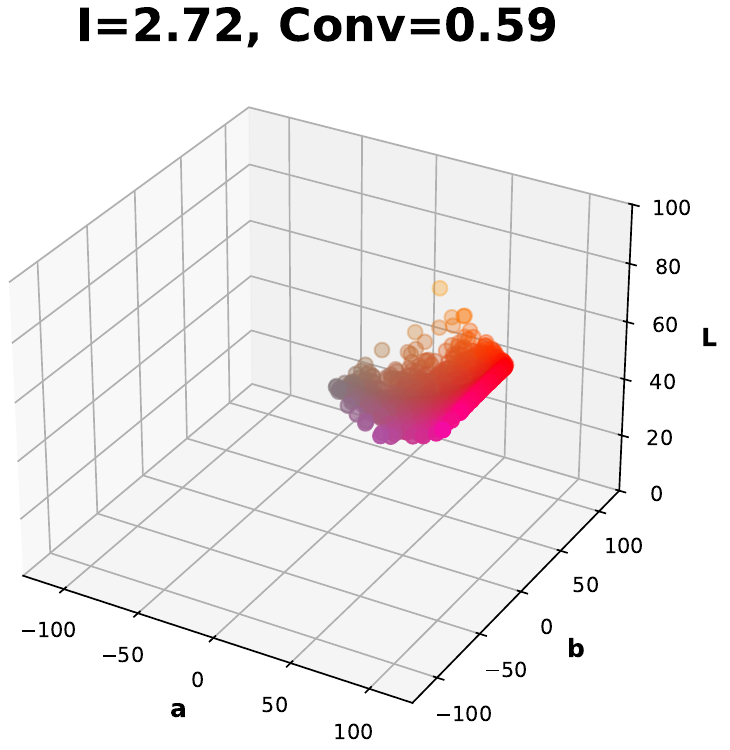}
  \caption{\textit{red}}
\end{subfigure}

\caption{Denotations of three color terms in CIELAB space under two training conditions
(top: 1 listener, no upsampling; bottom: 30 listeners, upsampling 100; both are after RL).}
\label{fig:convex_example}
\end{figure}

To illustrate the effect of the optimal condition, Figure \ref{fig:convex_example} compares denotation examples from the baseline setting (single listener, no upsampling) with those from the 30-listener, upsampling-100 setting. Under the latter, the denotations of most color terms, including both specific color names such as \textit{aqua} and more general terms like \textit{green} and \textit{red}, become more spatially compact and convex in CIELAB space. 

\vspace{.1cm}\noindent\textbf{Semantic Drift. }
We finally examine semantic drift, which measures how much the learned lexicon deviates from the human lexicon learned during SL. Importantly, higher semantic drift does not necessarily indicate poor agent performance; the representations may still be internally consistent and convex. As in other neural-agent-based simulations of language evolution \citep{lian-etal-2024-nellcom, zhang-etal-2024-endowing, lian2025simulating}, RL provides interactive communication settings, and semantic drift in RL provides a useful lens for exploring how different interaction conditions shape the properties of the emergent agent lexicon. 



Interacting with many listeners tends to increase the drift. Combined with the observation that more listeners increase convexity, this suggests that group communication amplifies pressure toward internally consistent but potentially less human-like semantic conventions. 
%
Upsampling under SL leads to a slight increase in drift, likely because a larger vocabulary introduces additional lexical distinctions.
Nevertheless, 
upsampling substantially reduces semantic drift in SL$+$RL conditions for a fixed listener setup.
For example, with one listener, $D_L$ decreases from 39.69 (no upsampling) to 35.11 (upsampling 100) and further to 27.11 (upsampling 200).
A similar decreasing trend is observed with 5 and 30 listeners. This pattern suggests that upsampling promotes more stable lexicons, making it more difficult for agents to shift denotations or establish idiosyncratic conceptual pacts over time. %

\vspace{.1cm}\noindent\textbf{Discussion. }
Overall, our results show that lexical diversity and informativeness are shaped by both the number of listeners and upsampling, with single-listener and moderate-to-high upsampling setups promoting richer and more informative lexicons. Convexity is generally high under SL, but RL combined with multiple listeners and moderate upsampling yields more human-like color representations. Semantic drift highlights the trade-off between alignment with human lexicons and internally consistent group communication. 

\section{Conclusion}

We investigated why neural agents trained in the NeLLCom-Lex framework develop lexicons that diverge from human color naming systems, despite achieving high communicative accuracy. 
Our experiments identify two factors that could influence the properties of emergent lexicons: frequency imbalance in the supervised training data and listener diversity during the referential game. Upsampling rare color terms increases lexical diversity, informativeness, and reduces semantic drift, while interacting with multiple listeners constrains idiosyncratic productions and promotes more convex semantic categories. Neither factor alone is sufficient; the most human-like lexicons emerge from their combination, balancing convexity, informativeness, vocabulary size, and drift. 

\section{Acknowledgments}
Arianna Bisazza is funded by the Talent Programme of the Dutch Research Council (NWO) under project VI.Vidi.221C.009. Gemma Boleda is funded by the Ministerio de Ciencia e Innovación and the Agencia Estatal de Investigación (Spain; grant PID2020-112602GB-I00/MICIN/AEI/10.13039/501100011033).

\printbibliography

@article{raviv2019larger,
  title={Larger communities create more systematic languages},
  author={Raviv, Limor and Meyer, Antje and Lev-Ari, Shiri},
  journal={Proceedings of the Royal Society B},
  volume={286},
  number={1907},
  pages={20191262},
  year={2019},
  publisher={The Royal Society}
}

@inproceedings{graesser2019emergent,
  title={Emergent linguistic phenomena in multi-agent communication games},
  author={Graesser, Laura Harding and Cho, Kyunghyun and Kiela, Douwe},
  booktitle={Proceedings of the 2019 conference on empirical methods in natural language processing and the 9th international joint conference on natural language processing (EMNLP-IJCNLP)},
  pages={3700--3710},
  year={2019}
}

@article{beckner2009language,
	title = {Language {Is} a {Complex} {Adaptive} {System}: {Position} {Paper}},
	volume = {59},
	issn = {1467-9922},
	url = {https://onlinelibrary.wiley.com/doi/abs/10.1111/j.1467-9922.2009.00533.x},
	doi = {10.1111/j.1467-9922.2009.00533.x},
	number = {s1},
	urldate = {2026-01-25},
	journal = {Language Learning},
	author = {Beckner, Clay and Blythe, Richard and Bybee, Joan and Christiansen, Morten H. and Croft, William and Ellis, Nick C. and Holland, John and Ke, Jinyun and Larsen-Freeman, Diane and Schoenemann, Tom},
	year = {2009},
	note = {\_eprint: https://onlinelibrary.wiley.com/doi/pdf/10.1111/j.1467-9922.2009.00533.x},
	pages = {1--26},
}

@article{chawla2002smote,
  title={SMOTE: synthetic minority over-sampling technique},
  author={Chawla, Nitesh V and Bowyer, Kevin W and Hall, Lawrence O and Kegelmeyer, W Philip},
  journal={Journal of artificial intelligence research},
  volume={16},
  pages={321--357},
  year={2002}
}

@inproceedings{henning2023survey,
  title={A survey of methods for addressing class imbalance in deep-learning based natural language processing},
  author={Henning, Sophie and Beluch, William and Fraser, Alexander and Friedrich, Annemarie},
  booktitle={Proceedings of the 17th Conference of the European Chapter of the Association for Computational Linguistics},
  pages={523--540},
  year={2023}
}

@article{ellis2002frequency,
  title={Frequency effects in language processing: A review with implications for theories of implicit and explicit language acquisition},
  author={Ellis, Nick C},
  journal={Studies in second language acquisition},
  volume={24},
  number={2},
  pages={143--188},
  year={2002},
  publisher={Cambridge University Press}
}

@inproceedings{zhang-etal-2025-nellcom,
    title = "{N}e{LLC}om-Lex: A Neural-agent Framework to Study the Interplay between Lexical Systems and Language Use",
    author = {Zhang, Yuqing  and
      {\"U}rker, Ecesu  and
      Verhoef, Tessa  and
      Boleda, Gemma  and
      Bisazza, Arianna},
    editor = "Christodoulopoulos, Christos  and
      Chakraborty, Tanmoy  and
      Rose, Carolyn  and
      Peng, Violet",
    booktitle = "Findings of the Association for Computational Linguistics: EMNLP 2025",
    month = nov,
    year = "2025",
    address = "Suzhou, China",
    publisher = "Association for Computational Linguistics",
    url = "https://aclanthology.org/2025.findings-emnlp.580/",
    doi = "10.18653/v1/2025.findings-emnlp.580",
    pages = "10929--10945",
    ISBN = "979-8-89176-335-7"
}

@article{steinert2020ease,
  title={Ease of learning explains semantic universals},
  author={Steinert-Threlkeld, Shane and Szymanik, Jakub},
  journal={Cognition},
  volume={195},
  pages={104076},
  year={2020},
  publisher={Elsevier}
}

@book{gardenfors2014geometry,
  title={The Geometry of Meaning: Semantics Based on Conceptual Spaces},
  author={G{\"a}rdenfors, Peter},
  year={2014},
  publisher={MIT Press}
}

@article{gardenfors2000conceptual,
  title={Conceptual Spaces: The Geometry of Thought},
  author={G{\"a}rdenfors, Peter},
  year={2000},
  publisher={The MIT Press}
}

@inproceedings{jager2010natural,
  title={Natural color categories are convex sets},
  author={J{\"a}ger, Gerhard},
  booktitle={Logic, Language and Meaning: 17th Amsterdam Colloquium, Amsterdam, The Netherlands, December 16-18, 2009, Revised Selected Papers},
  pages={11--20},
  organization={Springer},
  year={2009}
}

@article{jager2007evolution,
  title={The evolution of convex categories},
  author={J{\"a}ger, Gerhard},
  journal={Linguistics and philosophy},
  volume={30},
  number={5},
  pages={551--564},
  year={2007},
  publisher={Springer}
}

@book{hopper2003grammaticalization,
  title={Grammaticalization},
  author={Hopper, Paul J and Traugott, Elizabeth Closs},
  year={2003},
  publisher={Cambridge university press}
}

@article{tucker2025towards,
  title={Towards human-like emergent communication via utility, informativeness, and complexity},
  author={Tucker, Mycal and Shah, Julie and Levy, Roger and Zaslavsky, Noga},
  journal={Open Mind},
  volume={9},
  pages={418--451},
  year={2025},
  publisher={MIT Press 255 Main Street, 9th Floor, Cambridge, Massachusetts 02142, USA~…}
}

@article{zaslavsky2018efficient,
  title={Efficient compression in color naming and its evolution},
  author={Zaslavsky, Noga and Kemp, Charles and Regier, Terry and Tishby, Naftali},
  journal={Proceedings of the National Academy of Sciences},
  volume={115},
  number={31},
  pages={7937--7942},
  year={2018},
  publisher={National Academy of Sciences}
}

@article{li2019ease,
  title={Ease-of-teaching and language structure from emergent communication},
  author={Li, Fushan and Bowling, Michael},
  journal={Advances in neural information processing systems},
  volume={32},
  year={2019}
}

@article{pickering2004toward,
  title={Toward a mechanistic psychology of dialogue},
  author={Pickering, Martin J and Garrod, Simon},
  journal={Behavioral and brain sciences},
  volume={27},
  number={2},
  pages={169--190},
  year={2004},
  publisher={Cambridge University Press}
}

@article{ostrand2019repeat,
  title={Repeat after us: Syntactic alignment is not partner-specific},
  author={Ostrand, Rachel and Ferreira, Victor S},
  journal={Journal of memory and language},
  volume={108},
  pages={104037},
  year={2019},
  publisher={Elsevier}
}

@incollection{cangelosi2002computer,
  title={Computer simulation: A new scientific approach to the study of language evolution},
  author={Cangelosi, Angelo and Parisi, Domenico},
  booktitle={Simulating the evolution of language},
  pages={3--28},
  year={2002},
  publisher={Springer}
}

@book{hawkins2004efficiency,
  title={Efficiency and complexity in grammars},
  author={Hawkins, John A},
  year={2004},
  publisher={OUP Oxford}
}

@inproceedings{lian2025simulating,
 author = {Yuchen Lian and Arianna Bisazza and Tessa Verhoef},
 booktitle = {Proceedings of the 47th Annual Conference of the Cognitive Science Society (CogSci)},
 doi = {10.48550/arXiv.2502.04038},
 month = {July},
 title = {Simulating the Emergence of Differential Case Marking with Communicating Neural-Network Agents},
 url = {https://escholarship.org/uc/item/5dr7h5tp},
 year = {2025}
}

@incollection{de2006computer,
  title={Computer modelling as a tool for understanding language evolution},
  author={De Boer, Bart},
  booktitle={Evolutionary epistemology, language and culture: A non-adaptationist, systems theoretical approach},
  pages={381--406},
  year={2006},
  publisher={Springer}
}

@article{steels1997synthetic,
  title={The synthetic modeling of language origins},
  author={Steels, Luc},
  journal={Evolution of communication},
  volume={1},
  number={1},
  pages={1--34},
  year={1997},
  publisher={John Benjamins}
}

@article{lazaridou2020emergent,
  title={Emergent multi-agent communication in the deep learning era},
  author={Lazaridou, Angeliki and Baroni, Marco},
  journal={arXiv preprint arXiv:2006.02419},
  year={2020}
}

@book{campbell2013historical,
  title={Historical linguistics},
  author={Campbell, Lyle},
  year={2013},
  publisher={Edinburgh University Press}
}

@article{gualdoni_whats_2023,
	title = {What’s in a name? {A} large-scale computational study on how competition between names affects naming variation},
	volume = {133},
	journal = {Journal of Memory and Language},
	author = {Gualdoni, Eleonora and Brochhagen, Thomas and Mädebach, Andreas and Boleda, Gemma},
	year = {2023},
	note = {Publisher: Elsevier},
	pages = {104459},
}

@article{brochhagen2022languages,
  title={When do languages use the same word for different meanings? The Goldilocks principle in colexification},
  author={Brochhagen, Thomas and Boleda, Gemma},
  journal={Cognition},
  volume={226},
  pages={105179},
  year={2022},
  publisher={Elsevier}
}

@article{lian2023communication,
    author = {Lian, Yuchen and Bisazza, Arianna and Verhoef, Tessa},
    title = "{Communication Drives the Emergence of Language Universals in Neural Agents: Evidence from the Word-order/Case-marking Trade-off}",
    journal = {Transactions of the Association for Computational Linguistics},
    volume = {11},
    pages = {1033-1047},
    year = {2023},
    month = {08},
    issn = {2307-387X},
    doi = {10.1162/tacl_a_00587},
    url = {https://doi.org/10.1162/tacl\_a\_00587}
}

@inproceedings{lian-etal-2024-nellcom,
    title = "{N}e{LLC}om-{X}: A Comprehensive Neural-Agent Framework to Simulate Language Learning and Group Communication",
    author = "Lian, Yuchen  and
      Verhoef, Tessa  and
      Bisazza, Arianna",
    editor = "Barak, Libby  and
      Alikhani, Malihe",
    booktitle = "Proceedings of the 28th Conference on Computational Natural Language Learning",
    month = nov,
    year = "2024",
    address = "Miami, FL, USA",
    publisher = "Association for Computational Linguistics",
    url = "https://aclanthology.org/2024.conll-1.19/",
    doi = "10.18653/v1/2024.conll-1.19",
    pages = "243--258"
}

@inproceedings{kharitonov-etal-2019-egg,
    title = "{EGG}: a toolkit for research on Emergence of lan{G}uage in Games",
    author = "Kharitonov, Eugene  and
      Chaabouni, Rahma  and
      Bouchacourt, Diane  and
      Baroni, Marco",
    editor = "Pad{\'o}, Sebastian  and
      Huang, Ruihong",
    booktitle = "Proceedings of the 2019 Conference on Empirical Methods in Natural Language Processing and the 9th International Joint Conference on Natural Language Processing (EMNLP-IJCNLP): System Demonstrations",
    month = nov,
    year = "2019",
    address = "Hong Kong, China",
    publisher = "Association for Computational Linguistics",
    url = "https://aclanthology.org/D19-3010/",
    doi = "10.18653/v1/D19-3010",
    pages = "55--60"
}

@article{chaabouni2021communicating,
  title={Communicating artificial neural networks develop efficient color-naming systems},
  author={Chaabouni, Rahma and Kharitonov, Eugene and Dupoux, Emmanuel and Baroni, Marco},
  journal={Proceedings of the National Academy of Sciences},
  volume={118},
  number={12},
  pages={e2016569118},
  year={2021},
  publisher={National Academy of Sciences}
}

@article{monroe-etal-2017-colors,
    title = "Colors in Context: A Pragmatic Neural Model for Grounded Language Understanding",
    author = "Monroe, Will  and
      Hawkins, Robert X.D.  and
      Goodman, Noah D.  and
      Potts, Christopher",
    editor = "Lee, Lillian  and
      Johnson, Mark  and
      Toutanova, Kristina",
    journal = "Transactions of the Association for Computational Linguistics",
    volume = "5",
    year = "2017",
    address = "Cambridge, MA",
    publisher = "MIT Press",
    url = "https://aclanthology.org/Q17-1023/",
    doi = "10.1162/tacl_a_00064",
    pages = "325--338"
}

@inproceedings{gualdoni-boleda-2024-objects,
    title = "Why do objects have many names? A study on word informativeness in language use and lexical systems",
    author = "Gualdoni, Eleonora  and
      Boleda, Gemma",
    editor = "Al-Onaizan, Yaser  and
      Bansal, Mohit  and
      Chen, Yun-Nung",
    booktitle = "Proceedings of the 2024 Conference on Empirical Methods in Natural Language Processing",
    month = nov,
    year = "2024",
    address = "Miami, Florida, USA",
    publisher = "Association for Computational Linguistics",
    url = "https://aclanthology.org/2024.emnlp-main.1009/",
    doi = "10.18653/v1/2024.emnlp-main.1009",
    pages = "18150--18163"
}

@inproceedings{white2020learning,
  title={Learning to refer informatively by amortizing pragmatic reasoning},
  author={White, Julia and Mu, Jesse and Goodman, Noah D},
  booktitle={Proceedings of the Annual Meeting of the Cognitive Science Society},
  volume={42},
  year={2020}
}

@inproceedings{zhang-etal-2024-endowing,
    title = "Endowing Neural Language Learners with Human-like Biases: A Case Study on Dependency Length Minimization",
    author = "Zhang, Yuqing  and
      Verhoef, Tessa  and
      van Noord, Gertjan  and
      Bisazza, Arianna",
    editor = "Calzolari, Nicoletta  and
      Kan, Min-Yen  and
      Hoste, Veronique  and
      Lenci, Alessandro  and
      Sakti, Sakriani  and
      Xue, Nianwen",
    booktitle = "Proceedings of the 2024 Joint International Conference on Computational Linguistics, Language Resources and Evaluation (LREC-COLING 2024)",
    month = may,
    year = "2024",
    address = "Torino, Italia",
    publisher = "ELRA and ICCL",
    url = "https://aclanthology.org/2024.lrec-main.516/",
    pages = "5819--5832"
}

@article{carlsson2024cultural,
  title={Cultural evolution via iterated learning and communication explains efficient color naming systems},
  author={Carlsson, Emil and Dubhashi, Devdatt and Regier, Terry},
  journal={Journal of Language Evolution},
  volume={9},
  number={1-2},
  pages={49--66},
  year={2024},
  publisher={Oxford University Press UK}
}

@incollection{brainard2003color,
  author    = {David H. Brainard},
  title     = {Color appearance and color difference specification},
  booktitle = {The Science of Color},
  edition   = {2},
  pages     = {191--216},
  year      = {2003},
  publisher = {Elsevier},
  address   = {Oxford}
}

@article{regier2015word,
  title={Word meanings across languages support efficient communication},
  author={Regier, Terry and Kemp, Charles and Kay, Paul},
  journal={The handbook of language emergence},
  pages={237--263},
  year={2015},
  publisher={Wiley Online Library}
}

@article{Carlsson2020,
    doi = {10.1371/journal.pone.0234894},
    author = {Kågebäck, Mikael AND Carlsson, Emil AND Dubhashi, Devdatt AND Sayeed, Asad},
    journal = {PLOS ONE},
    publisher = {Public Library of Science},
    title = {A reinforcement-learning approach to efficient communication},
    year = {2020},
    month = {07},
    volume = {15},
    url = {https://doi.org/10.1371/journal.pone.0234894},
    pages = {1-26},
    number = {7},

}

@article{loreto2012origin,
  title={On the origin of the hierarchy of color names},
  author={Loreto, Vittorio and Mukherjee, Animesh and Tria, Francesca},
  journal={Proceedings of the National Academy of Sciences},
  volume={109},
  number={18},
  pages={6819--6824},
  year={2012},
  publisher={National Academy of Sciences}
}

@article{steels2005coordinating,
  title={Coordinating perceptually grounded categories through language: A case study for colour},
  author={Steels, Luc and Belpaeme, Tony and others},
  journal={Behavioral and brain sciences},
  volume={28},
  number={4},
  pages={469--488},
  year={2005},
  publisher={[New York]: Cambridge University Press, 1978-}
}

@inproceedings{bleys2009grounded,
  title={The grounded colour naming game},
  author={Bleys, Joris and Loetzsch, Martin and Spranger, Michael and Steels, Luc},
  year={2009},
  booktitle={Proceedings of the 18th IEEE International Symposium on Robot and Human Interactive Communication (RoMan 2009)}
}

@article{
Baronchelli2010,
author = {Andrea Baronchelli  and Tao Gong  and Andrea Puglisi  and Vittorio Loreto },
title = {Modeling the emergence of universality in color naming patterns},
journal = {Proceedings of the National Academy of Sciences},
volume = {107},
number = {6},
pages = {2403-2407},
year = {2010},
doi = {10.1073/pnas.0908533107},
URL = {https://www.pnas.org/doi/abs/10.1073/pnas.0908533107},
eprint = {https://www.pnas.org/doi/pdf/10.1073/pnas.0908533107}}

@incollection{cook2005world,
  title={The world color survey database},
  author={Cook, Richard S and Kay, Paul and Regier, Terry},
  booktitle={Handbook of categorization in cognitive science},
  pages={223--241},
  year={2005},
  publisher={Elsevier}
}

@book{berlin1969basic,
  title={Basic color terms: Their universality and evolution},
  author={Berlin, Brent and Kay, Paul},
  year={1991},
  publisher={University of California Press, Berkeley}
}

@book{clark1996using,
  title={Using language},
  author={Clark, Herbert H},
  year={1996},
  publisher={Cambridge university press}
}

@article{brennan1996conceptual,
  title={Conceptual pacts and lexical choice in conversation},
  author={Brennan, Susan E and Clark, Herbert H},
  journal={Journal of experimental psychology: Learning, memory, and cognition},
  volume={22},
  number={6},
  pages={1482},
  year={1996},
  publisher={American Psychological Association}
}

@phdthesis{brochhagen2018signaling,
  title={Signaling under uncertainty},
  author={Brochhagen, Thomas},
  year={2018},
  school={University of Amsterdam}
}

@article{steels1998spatially,
  title={Spatially distributed naming games},
  author={Steels, Luc and McIntyre, Angus},
  journal={Advances in complex systems},
  volume={1},
  number={04},
  pages={301--323},
  year={1998},
  publisher={World Scientific}
}

@article{tieleman2019shaping,
  title={Shaping representations through communication: community size effect in artificial learning systems},
  author={Tieleman, Olivier and Lazaridou, Angeliki and Mourad, Shibl and Blundell, Charles and Precup, Doina},
  journal={arXiv preprint arXiv:1912.06208},
  year={2019}
}

@inproceedings{michelrevisiting,
  title={Revisiting Populations in multi-agent Communication},
  author={Michel, Paul and Rita, Mathieu and Mathewson, Kory Wallace and Tieleman, Olivier and Lazaridou, Angeliki},
  year={2023},
  booktitle={The Eleventh International Conference on Learning Representations}
}

\end{document}